# Finding Numerical Solutions of Diophantine Equations using Ant Colony Optimization

Siby Abraham*, Sugata Sanyal[a], Mukund Sanglikar[b]

Department of Maths & Stats, Guru Nanak Khalsa College, University of Mumbai, India
[a] Corporate Technology Organization, Tata Consultancy Services, Gateway Park, Andheri (E), Mumbai, India
[b] Department of Mathematics, Mithibai College, University of Mumbai, India.

**Abstract**

The paper attempts to find numerical solutions of Diophantine equations, a challenging problem as there are no general methods to find solutions of such equations. It uses the metaphor of foraging habits of real ants. The ant colony optimization based procedure starts with randomly assigned locations to a fixed number of artificial ants. Depending upon the quality of these positions, ants deposit pheromone at the nodes. A successor node is selected from the topological neighborhood of each of the nodes based on this stochastic pheromone deposit. If an ant bumps into an already encountered node, the pheromone is updated correspondingly. A suitably defined pheromone evaporation strategy guarantees that premature convergence does not take place. The experimental results, which compares with those of other machine intelligence techniques, validate the effectiveness of the proposed method.

**Keywords:** Ant Colony Optimization, Diophantine equation, Pheromone

## 1 Introduction

Ant colony optimization (ACO) [14, 16, 17] is a population based optimization technique, which mimics social behaviour of real ants [8]. Investigations to social behaviour of insects, initiated by Pierre-Paul Grasse, reveal that ants deposit on the ground a substance called pheromone while moving from food sources to the nest and vice versa. The deposits of pheromone by ants create a pheromone trail on the path. Other ants can smell this pheromone and hence can be influenced by the path having high pheromone concentrations [15].

In a computational perspective, one could look at this as a distributed optimization mechanism in which small contributions offered by an ensemble of ants result in an emergent behaviour called finding an optimal path. Thus, finding solution of an optimization problem can be visualised as a structured movement of ants on a weighted graph building solutions incrementally. The incremental construction of possible solutions follows a stochastic model based on deposit and evaporation of pheromone. This is realised by a set of parameters associated with nodes and edges, which are changed at run time by the movement of ants.

ACO is used as an optimization technique in many different fields. The primary problem areas tackled by ACO are combinatorial optimization problems, which include problems like routing, scheduling, subset finding, assignment and layout, machine learning and bioinformatics. ACO also applies in multi objective optimization, stochastic optimization, dynamic optimization, continuous optimization, communication networks and many industrial problems [36].

Though ACO has been applied in different fields, there are few applications in the domain of Pure Mathematics. This paper is an attempt to apply ACO in the field of pure Mathematics. The work is significant as it attempts to find numerical solutions of a very challenging Mathematical problem. The problem is referred as 'challenging' because, though it is historically important and has many applications in many fields, there are no formal methods to find solutions of this problem. The rest of the paper is divided into different sections. Section 2 gives detailed introduction of Diophantine equations where the ACO technique is applied. Section 3 explains the methodology used. Section 4 deals with the experimental results and section 5 presents the conclusion.

## 2 Diophantine Equations

A Diophantine equation [11] is a polynomial equation, given by

$$f(a_1, a_2, \ldots, a_n, x_1, x_2, \ldots, x_n) = N \ldots\ldots\ldots(1)$$

where $a_i$ and N are integers.

---

* Corresponding author: Siby Abraham, email:sibyam@gmail.com, tel:(+91)2224096234

The equation (1) may have:

- No nontrivial, integral solution as in the case of $x_1^2 + x_2^2 + x_3^2 = 0$.
- Finitely many integral solutions as in the case of $x_1^4 + x_2^4 - x_3^4 = 0$.
- Infinitely many integral solutions as in the case of $x_1^n + x_2^n - x_3^n = 0$.

Diophantine equations and their particular types has been an object of interest for many generations of Mathematicians [9, 10]. This inspired numerous attempts to find solutions of special cases of such equations. For example, linear equation

$$a_1 x_1 + a_2 x_2 + \ldots\ldots + a_n x_n = N$$

has a solution if and only if the greatest common divisor $(a_1, a_2, \ldots, a_n)$ divides N. The equation

$$x_1^n + x_2^n - x_3^n = 0$$

is the equation in Fermat's last theorem [18, 34] which has no integral, non-trivial solutions for n>2. For n = 2, the equation reduces to

$$x_1^2 + x_2^2 - x_3^2 = 0$$

whose solutions are known as Pythagorean triplets and are primitive in nature. There is at least one integral solution to the equation

$$x_1^2 + x_2^2 + x_3^2 + x_4^2 = N,$$

which Waring conjectured that they can be generalized to higher powers with 9 cubes and 19 fourth powers [41].

These individual and isolated attempts to find solutions of particular types of Diophantine equations have received an impetus by Hilbert. He, in his famous tenth problem [20], posed the question that whether there does exist an algorithm to find solutions of general Diophantine equations with integral coefficients. Decades later, Matiyasevich [28, 13] conclusively proved that it is impossible to obtain a general solution.

This has not dampened interests in Diophantine equations, as the scope and importance of Diophantine equations are not restricted to the abstract realm of Number theory. It has applications in many fields like Public key cryptosystems [26, 27], Data dependency in Super Computers [37, 40], Integer factorization, Algebraic curves [31], Projective curves [11, 35], Elliptic curves [25], Computable economics [38] and Theoretical Computer Science [23, 19].

In this context, finding numerical solutions of Diophantine equations is considered as the only possible way out considering the importance of the field. However, this turns out to be a challenging problem taking into account the computational complexity involved. One can estimate that the search space [33] of the equation (1) consists of $N^n$ possible solutions, provided $x_i$'s are allowed to take only positive integers. Literature mentions few attempts to apply traditional search techniques in this area. More common and standard techniques like Breadth First Search (BFS) and Depth First Search (DFS) are of not much use as BFS encounters space and time complexities whereas DFS can get trapped in blindly following a non- solution. Though A* search was used by Abraham and Sanglikar [6], it was found that the system ran out of space fast especially in the case of large equations. It has been observed that soft computing techniques are much more effective because of its proven ability to manoeuvre huge state space. Hsiung and Mathews [22] used a first-degree linear Diophantine equation to illustrate the basic concepts of a Genetic Algorithm (GA). Abraham and Sanglikar [1] tried Genetic Algorithm formally to find numerical solutions of Diophantine equations by applying genetic operators-mutation and crossover [29]. Though the procedure could find solutions for smaller equations, it was not fully random in nature and seemed more like a steepest ascent hill climbing and hits on local optimum points. Abraham et al [2] and Abraham and Sanglikar[3] proposed a novel reciprocally induced co-evolution method [30] [32] based on 'host parasite co-evolution' [21, 29, 39] to tide over the repeated occurrence of local hilltops in a typical GA. Joya et al [24] applied higher order Hopfield neural network and Abraham et al [4] used feed forward back propagation network to find numerical solutions of Diophantine equation. Abraham and Sanglikar [5] proposed simulated annealing as a viable search strategy for finding numerical solutions of a Diophantine equation. Abraham et al [7] used Particle Swarm Optimization (PSO) as a swarm intelligence tool to find the numerical solutions and compared the results obtained with other artificial intelligence techniques driving home the effectiveness of swarm intelligence in this domain. The present work, which uses Ant Colony Optimization, is inspired by the positive results obtained by PSO.

# 3 ANT-DOES

The system developed to find numerical solutions of Diophantine equations using Ant Colony Optimization (ACO) is referred as Ant Colony Optimization based Diophantine Equation Solver (ANT-DOES). The procedure, whose pseudo code is given in Figure 1 (which is explained in detail in sections 3.1, 3.2 and 3.3), uses a particular type of equation as given below for the description and experimental purpose

$$a_1 x_1^{p_1} + a_2 x_2^{p_2} + \ldots\ldots + a_n x_n^{p_n} = N \quad \ldots\ldots (2)$$

where $a_i \in Z$, the set of all integers and $p_i \in Z^+$, $x_i \in Z^+$, where $Z^+$ is the set of all positive integers, for i = 1, 2, 3, ..., n and n ∈ N. Other types of Diophantine equations can also be treated in similar fashion. However, the experimental results discussed in the remaining part of the paper covers only Diophantine equations as given in equation (2).

## 3.1 Initialization of population of nodes

The ANT-DOES procedure starts with creation of a population of a fixed number of initial nodes from where the search process is initiated. Each node is an n- dimensional vector given by $(x_1, x_2, \ldots, x_n)$. Each coordinate of this vector is a randomly generated positive integer between 1 and $p = \text{int}\{(N^{(1/\text{minPower})}\} + 1$ where 'minPower' is the minimum of all powers of the given equation. By using 'minPower', instead of any other heuristic, the procedure tries to encompass availability of all possible solutions in the search. This is because, the range of the coordinates of a possible solution lies between 1 and 'int$\{(N^{(1/\text{minPower})}\} + 1$'. Also, such a heuristic reduces substantially the complexity of search space from $N^n$ possible nodes to a better manageable level.

ANT-DOES create that many numbers of fixed artificial ants as that of the initial nodes generated. These ants reside on these nodes. The procedure involves movement of these ants through different nodes in subsequent steps.

## 3.2 Fitness of nodes

The effectiveness of generated nodes is tested using a fitness function. For the node $x = (x_1, x_2, \ldots, x_n)$, it is given by

$$f(x) = \text{Abs}( N - (a_1 x_1^{p_1} + a_2 x_2^{p_2} + \ldots\ldots + a_n x_n^{p_n})) \quad \ldots (3)$$

A node whose fitness value, which is always taken as positive, equals to zero is taken as a solution. The value of fitness gives an idea about how far the node is from the solution of the equation. The guiding spirit behind the search process is to find nodes, which minimizes the value of this fitness function through movement of ants.

## 3.3 Creation of neighbors

The neighbors of a node are chosen using a special strategy. As per this, for each node, a fixed number of neighbors are generated within an admissible range. The coordinates of each of these neighbors are altered by adding a random number between 1 and $p = \text{int}\{(N^{(1/\text{minPower})}\} + 1$ to the value of the coordinate of the node. If the updated value of the coordinate is within the admissible range of the nodes, we retain the value. If this number crosses the range, we take the remainder on dividing the number by 'p'. Similarly, other coordinates of the neighbor are constructed. That means, the neighbours of $(x_1, x_2, \ldots, x_i, \ldots x_n)$ are the collection of nodes $(x_{p1}, x_{p2}, \ldots, x_{pi}, \ldots, x_{pn})$ where each $x_{pi}$ can be at a maximum distance of modulo 'p' from the coordinate $x_i$.

So, formally, $x_{pi}$ is chosen randomly from the topological neighborhood given by equation (4)

$$Zx_i = \{x_{pi} \in Z^+ \text{ such that } |x_i - x_{pi}| < \text{modulo p} \} \quad \text{---} \quad (4)$$

Thus, each of the coordinate of the neighbour is within the distance of modulus 'p'. So, the neighbors of a node is a collection of randomly generated fixed number of nodes of type say $(x_1, x_2, \ldots, x_n)$ within the maximum distance of modulo p, where

$$p = \text{int}\{(N^{(1/\text{minPower})}\} + 1 \quad \text{---} \quad (5)$$

The creation of neighbours is illustrated using an example here. Consider an elementary Diophantine equation $x_1^2 + x_2^3 = 108$. Here minPow = min {2, 3} = 2. Therefore, the values of the coordinates of random initial nodes will lie between 1 and int $(108^{1/2}) = 10$. Let one of the nodes be (5, 6). Suppose we are interested in the neighbors of this node: For the first coordinate '5' of (5, 6), we generate a random number between 1 and 10. Suppose it is 3. Then this 3 is added to 5 giving the first coordinate as 8, which is less than p = 10. So, this value is accepted as it is. If the generated number is say 8, the same is added to 5 to get 13. Since 13 is greater than p = 10, we take the coordinate of the neighbor as 13 % 10 = 3. Thus, the first coordinate of the neighbor of (5, 6) is chosen as 3. Similar strategy is followed for finding each coordinate of the neighbor.

This procedure ensures that each node, where an ant resides, generates a fixed number of neighbors. This creation of neighbors within the distance of modulo 'p' for each of the node guarantees a steady set of possible candidate solutions. Each ant chooses a node within its topological neighborhood to move forward based on the fitness values of the neighbouring nodes. This selection of fitter nodes from the topological neighborhood by each ant completes the search procedure.

### 3.3 Pheromone model

The effectiveness of a better node with respect to other nodes in the neighborhood is given preference using a pheromone model. The pheromone model used in the work is slightly differed from the way it is practiced in other minimum path problems like Travelling Salesman problem. Usually pheromone is considered to be deposited by ants on the edges, the path connecting the nodes. The rationale of such a procedure is to identify the path having maximum pheromone deposits as the one frequently travelled by a larger number of ants, which can act as a candidate for an optimal path for the solution. In such cases, the objective of the problem is to find an optimal path. However, finding numerical solutions of Diophantine equations is not an optimal path problem as the objective is to find a solution node. So, we don't give much importance to the edges connecting nodes visited by the ants. Here, instead of pheromone being considered to be deposited on the edges, as is the usual practice, the pheromone is allowed to be deposited at the nodes. A pheromone distribution scheme, as is discussed in the section 3.3.1, is followed for a node where an ant resides. The pheromone content is assumed to be zero for a node, which has not been encountered by an ant.

**3.3.1 Pheromone Distribution Scheme:** As per this, an ant deposits a pheromone amount equivalent to its effectiveness of the node where it resides. Based on the strength of the pheromone deposit at the node, the surrounding ants could be attracted to choose the node with higher pheromone content. The quantity of pheromone deposited by an ant at a visited node is given by

$$Phero(x) = 1 / f(x) \quad \text{------} \quad (6)$$

where $f(x)$ is the fitness of the node x where the ant is located now. This simple and straight forward depiction of pheromone is derived from the fact that the present problem is a minimisation problem to find a node with fitness value zero. Thus, nodes with smaller fitness values have been given greater pheromone content as they are close to the solutions. This way, more ants are encouraged to traverse through such nodes.

Once each of the nodes in the topological neighborhood of a node has been allocated the pheromone, the ant chooses a node using the probability

$$Prob(x) = Phero(x) / \sum Phero(x) \quad \text{....} \quad (7)$$

Here the summation applies to all nodes, which are in the topological neighborhood of the given node. In practice, this is implemented by generating a uniform random number by the system and selecting the node satisfying the probability condition.

**3.3.2 Pheromone update:** The nodes which have already been encountered in the search process have been given additional weightage in the search by offering an additional amount of pheromone. The pheromone update used in the work is given by equation (8):

$$PheroUpdate(x) = 0.01 * Phero(x) \quad \text{...} \quad (8)$$

This additional pheromone content is offered in addition to the existing pheromone for the nodes, which have already been visited by another ant.

**3.3.3 Pheromone evaporation:** The unchecked and continuous update of pheromone at the nodes, which have already been visited by other ants, creates a possibility of premature convergence, the phenomenon in which most of the ants are forced to visit the nodes with an accumulated pheromone deposit. This problem is tackled in ANT-DOES by a pheromone evaporation strategy. As per this, the pheromone content is reduced gradually when an ant visits a node, which has already been visited more than once by other ants. The procedure adopts the following formula for the pheromone evaporation

$$PheroEvaporation(x) = (\text{Number of encountered visits} * Phero(x))/100 \quad \text{.....} \quad (9)$$

In addition to this, when an ant gets stuck on a local minimum point (because of not having a better neighbor), the pheromone content of that node gets completely evaporated. Then, the ant is forced to go back to the previous node to take another route.

### 3.4 Construction of graph

The whole process of movement of ants in the search space is akin to construction of a graph (x, e) where 'x' is a node and 'e' is an edge connecting the node 'x' with another node. The solution of the problem is reached when the construction graph reaches a node x with fitness f(x) equals zero. At every stage, the construction of the graph is iteratively directed towards the solution node by creating neighboring nodes in the vicinity of the current node. The selection of a node in the topological neighborhood of the current node is based on the probability given by equation (7), based on pheromone deposit given by equation (7), which is updated using equation (8) and equation (9). Thus, the stochastic selection of a node from the topological neighborhood is based on the configuration of the node 'x', its neighborhood structure and on the number prior visits of that node by the ants.

The selected node is now added to the graph and the process is iteratively continued. This iterative construction of the graph by creating neighbors by considering effectiveness of a node, its neighborhood structure and the prior encounters by ants is significant as this process differs from the way ACO algorithms are applied in typical, usual manner. This unique way of generating nodes in the neighborhood offers many candidate solutions in the search process. This helps in choosing a better node from a collection of candidate solutions, which accelerates the process of settling down to a solution much faster.

### 3.5 Sequence of solutions

ANT-DOES not only offer one solution of a given Diophantine Equation, but it offers many solutions in the search process. Once the user opts for more solutions, the procedure is reinitialised. The whole process is continued with a new random initial allocation of nodes for ants. This results in the search procedure trying new and untried paths in the search space forcing to find new solutions on the way. Thus, the procedure guarantees to have as many numerical solutions of a given equation as possible. The generation of many numerical solutions of a given equation is significant as there is no general way to find solutions of a general Diophantine Equation.

### 3.6 Termination condition

The procedure is temporarily terminated when a solution is reached. When a solution is found, the system offers to have more solutions, thereby looking for all possible sets of solutions. The procedure is halted when the number of iterations specified is exhausted.

## 4 Experimental results and discussion

The methods discussed in the paper were implemented in Java. The different nodes taken by an ant were represented by arrays. The movement of ants and pheromone trail were also implemented using arrays. The result of running ANT-DOES strategy reveals some important characteristics of ant colony optimization in addition to finding solutions of the given Diophantine equation. The following results were obtained by running the system with 10 ants with 10 neighbors for each ant, if not specified otherwise.

### 4.1 Equations with different powers and different number of variables

The system was run on Diophantine equations of different types. Table 1 shows the results obtained for eleven equations with different number of variables. Table 2 demonstrates the effectiveness of running the system with thirteen different Diophantine equations whose powers are different. The experimental results reveal that the system offers solutions of Diophantine equations with reasonably high values of powers and number of variables. It reveals that the procedure is effective in providing solutions where coordinates are distinctly placed. It also demonstrates that solutions of complex equations can also be obtained by providing enough computing power and time.

### 4.2 Generation of Different Solutions for an Equation

The system not only generates a solution for a given Diophantine equation, but it offers as much number of solutions as possible. Table 3 illustrates this using an elementary equation $x_1^2 + x_2^2 + x_3^2 = 2445$ as an example. It gives a list of first ten solutions obtained and the number of iterations consumed for the purpose. The second column of the table shows that the solutions are distinct and variable. That means, the system could be able to deliver solutions which look entirely different in the subsequent iterations.

### 4. 3 Position of ants during the search

Figure 2 and Figure 3 show positions of ants at the initial and at the end of the search process of finding solution. The result is obtained by running the system using an elementary equation $x_1^2 + x_2^2 = 9000$ for demonstration purpose, the results of which can be extrapolated for other equations also. Though on the first look both the figures look almost same, they tell different stories. At the beginning, as Figure 2 shows, the positions of ants are at random positions. As the process becomes more directed and mature, the ants take strategic locations, which are shown in Figure 3. The positions at this stage are where the ants are closer or moving closer to the solution node, which is (54, 78). It is to be noted that closeness does not mean that the nodes are physically closer, but they are closer to the solution in terms of better (means closer to zero) fitness values. As the ants come closer to the solution, they don't take any drastic movements as in the case of initial stages. They take positions, which are very near to the current best positions. This slow movement towards the end of the search helps in finding the solution without actually slipping out of the search process. This is because as there is randomness at every stage of the movement, sudden and drastic movement can force an ant to skip a solution within the vicinity of the current position.

### 4. 4 Pheromone deposits

Figure 4 and Figure 5 illustrate the pheromone deposits at the nodes by the ants at the initial stage and at the end of the process respectively. The equation used for the purpose is the same as discussed above i.e $x_1^2 + x_2^2 = 9000$. The pheromone deposits at the beginning show that they are smaller in values. This is due to the random positions taken by the ants and the low fitness values at the beginning. As the procedure gets stabilised, the amount of pheromone deposit show a big difference. The pheromone content becomes larger in quantity.

### 4.5 Number of ants and iterations:

Figure 6 shows the number ants used and the iterations required to find the first solution. An elementary Diophantine equation $x_1^2 + x_2^2 = 10125$ with five neighbors for each node was used for experimental purpose. When the number of ants was a small number like five, it required a large number of iterations to find a solution. However, it has been noted that just by increasing the number of ants, the solutions were not found fast. Around ten to twenty five ants at a time are sufficient to get a solution of a Diophantine equation within a reasonable number of iterations.

### 4.6 Number of neighbors and iterations:

Figure 7 indicates the relation between number of neighbors of a node and the number of iterations required to find the first solution. The equation $x_1^2 + 2 x_2^2 = 5400$ and ten ants were used in the experiment. It shows that the indiscriminate increase of the number of neighbors does not just result in a faster solution. It was observed that around 10 neighbors per node is a sufficient number to get a solution.

### 4.7 Comparative study with other techniques:

The table 4 gives a comparative study of the different techniques used to find numerical solutions of Diophantine Equations. The study covers six different techniques in addition to ANT-DOES, all of which were tried to find numerical solutions of Diophantine equations. The techniques used are Breadth First Search (BFS), Depth First Search (DFS), A* Search, Hill Climbing, Genetic Algorithm (GA), Particle Swarm Optimization (PSO) and ANT-DOES. Twelve different parameters or features were used in this comparative study. A brief description about the important features is given below:

- Representation: ANT-DOES represents candidate solutions as vectors as in the case of BFS, DFS, A* and Hill Climbing whereas GA represent them as chromosomes or strings. PSO depicts the candidate solutions as locations of particles.
- Computational style: The computational style of ANT-DOES is parallel as is the case with that of GA and PSO. All other techniques follow sequential styles.
- Procedure type: The procedure followed is random for ANT-DOES, GA and PSO and whereas it is deterministic for all other methods.
- Mathematical structure: ANT-DOES and PSO use graphs as formal mathematical structures whereas other techniques except GA follow tree formats.
- Formation of new states: New states are formed in ANT-DOES, GA and PSO stochastically while others follow production rules.
- Occurrence of local optima: ANT-DOES tackles the occurrence of local optimum points using judiciously applied pheromone evaporation strategy whereas GA needs some outside technique like 'host parasite coevolution' mechanism to tide over local optima. PSO does not report the occurrence of local optimum points.
- Convergence to Solution: ANT-DOES also follow the style of rapid movement initially and then a steady flow as in other effective techniques like GA and PSO.
- Appearance as solution: ANT-DOES express the solution as position of the ants. The GA shows it as evolved chromosome where as PSO represent solutions as evolved particles. All other techniques show them as nodes.

- Coordinates of solutions: Only GA, PSO and ANT-DOES could offer solutions whose coordinates are distinct.

A Diophantine equation can have many solutions. The techniques other than GA, PSO and ANT-DOES fail to give solutions whose coordinates are distinct. However, GA requires some other technique to overcome the problem of the occurrence of local optimum points. Also, GA is not successful to the extent of PSO in finding solutions of large and bigger Diophantine equations. Though ANT-DOES encounters local optimum points, it can overcome them by inbuilt methodology of pheromone evaporation, effectively giving comparable efficiency of that of PSO.

### 4.8 Similar nature of solutions offered by PSO and ANT-DOES

ANT-DOES follow a similar strategy and offer results similar to that offered by PSO. A closer look at the experimental results obtained by ANT-DOES and PSO for comparatively complex equations involving many variables gives a better insight. Table 5 shows the first solutions obtained by PSO and ANT-DOES for different equations. It is to be noted that both the methods used similar parameters. While ANT-DOES used ten ants for the experiment and PSO used ten particles in the run up. Fitness function used, heuristic adopted and stopping criterion used were also the same. As the table shows both could offer solutions of comparable nature, though we could not guarantee the same first solution by both the techniques.

### 4.9 Number of iterations used by PSO and ANT-DOES

Figure 8 shows the number of iterations used by PSO and ANT-DOES for getting the first solutions given in Table 5. It conveys that ANT-DOES offer comparable if not faster solutions than PSO. Thus, ant colony optimization can be credited as a viable strategy to offer numerical solutions of Diophantine solutions in an effective and faster way in comparison with other techniques used.

## 5 Conclusion

The paper presents a novel methodology involving ant colony optimization to find numerical solutions of Diophantine equations. The procedure, initiated with randomly generated positions for a fixed number of ants, is guided by pheromone deposits and heuristic information. The procedure has been validated with different Diophantine equations of varying degrees and powers. It is compared with other techniques to show its effectiveness as a better technique.

The further work involves enlarging the scope of the work to take care of highly complex equations with large number of variables and large powers by properly updating the pheromone strategy followed. It is expected that a better pheromone deposit and evaporation strategy will fit the requirement.

# List of Figures

```
Begin
        Initialize population of nodes;
        Initialize ant positions;
                While (solution node not reached)
                        For each ant
                                Create neighbors;
                                Find fitness;
                                Deposit pheromone;
                        End for
                        Apply pheromone update;
                        Apply Pheromone Evaporation;
                End while
End.
```

Figure 1. Pseudo code of ACO-DOES

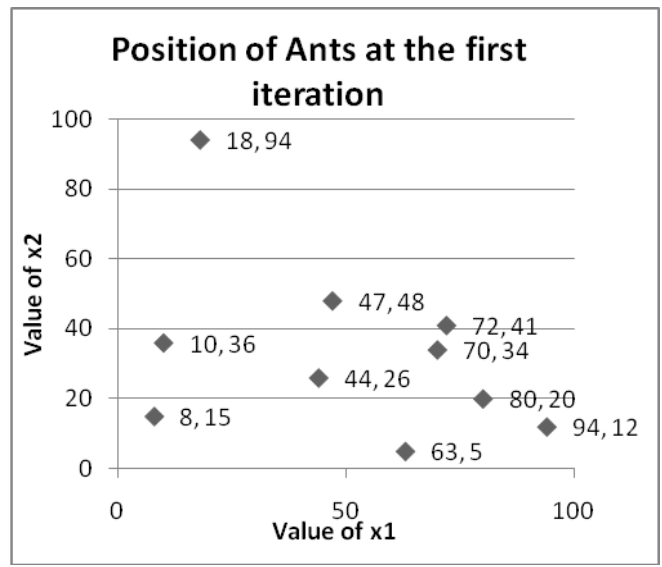

Figure 2: Positions of ants at the starting of the search process

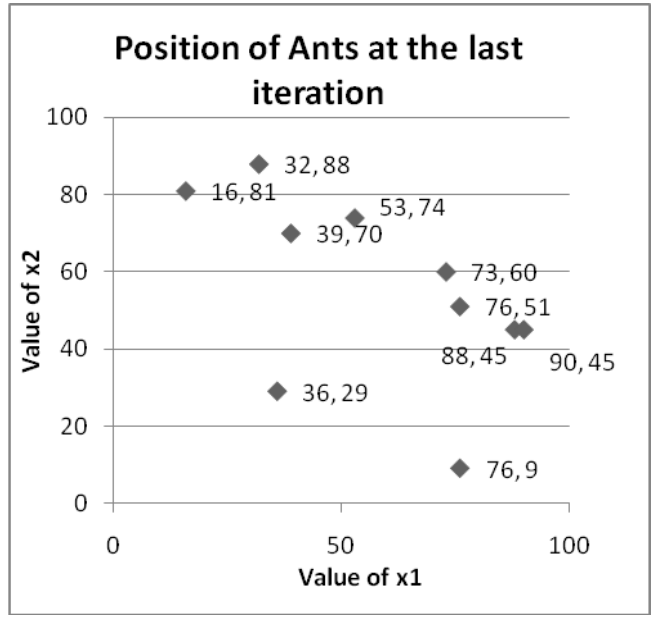

Figure 3: Positions of ants at the end of the search process

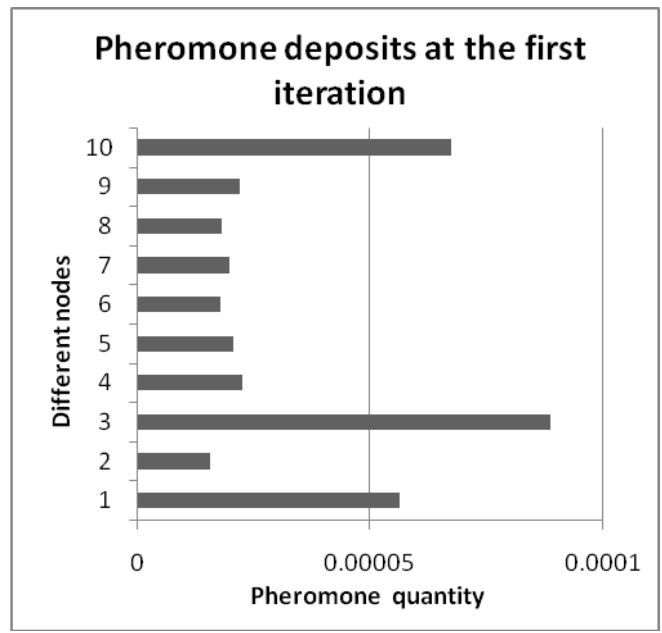

Figure 4: Distribution of Pheromone deposits at the beginning of search

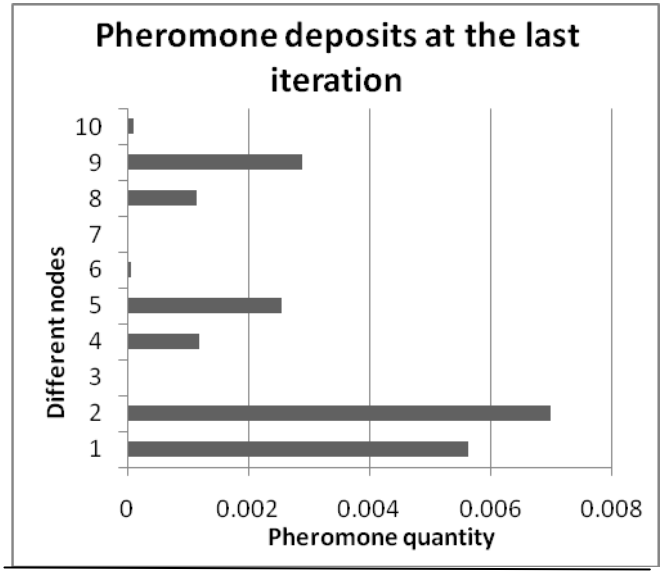

Figure 5: Distribution of Pheromone deposits when the solution is reached

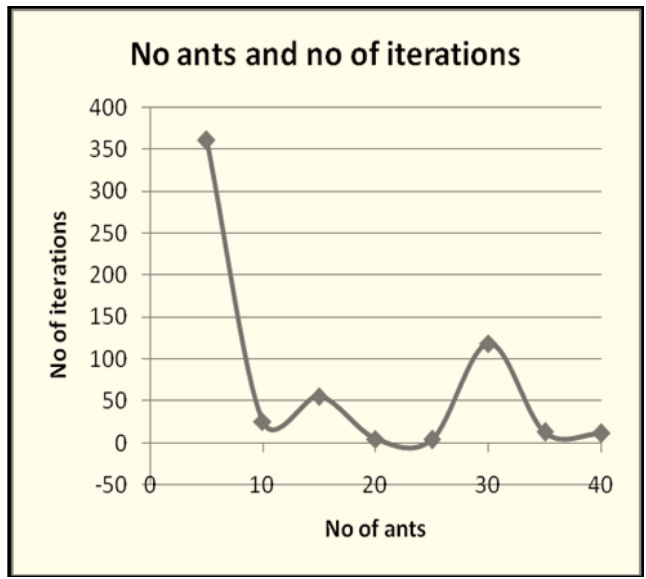

Figure 6: Number of ants and number of iterations required to find a solution

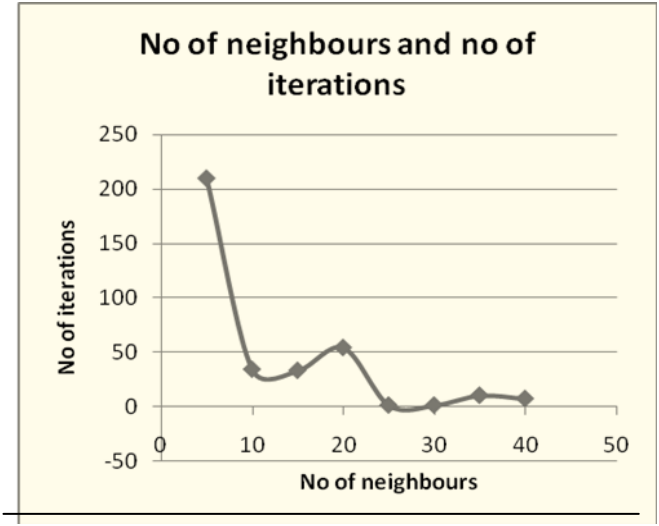

Figure 7: Number of neighbors of ants and number of iterations required to find a solution

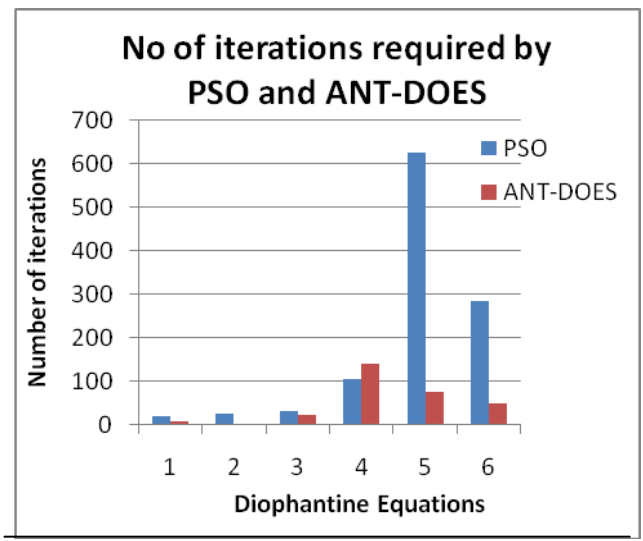

Figure 8: Number of iterations used by PSO and ANT-DOES to find a solution